\documentclass[final]{cvpr}

\usepackage{times}
\usepackage{epsfig}
\usepackage{graphicx}
\usepackage{amsmath}
\usepackage{amssymb}

\usepackage{subcaption}
\usepackage{tabularx}
\usepackage{booktabs}
\usepackage{siunitx}
\usepackage{dsfont}
\usepackage{verbatim}
\usepackage{placeins}

\usepackage[pagebackref=true,breaklinks=true,colorlinks,bookmarks=false]{hyperref}

\def\cvprPaperID{16} 
\def\confYear{CVPR 2021}

\newcommand*{\arxiv}{}%

\begin{document}

\title{DeepShift: Towards Multiplication-Less Neural Networks}

\author{Mostafa Elhoushi\textsuperscript{1}, Zihao Chen\textsuperscript{1,2}, Farhan Shafiq\textsuperscript{1}, Ye Henry Tian\textsuperscript{1}, Joey Yiwei Li\textsuperscript{1}\\
\textsuperscript{1}Huawei Technologies, Markham, Canada\\
\textsuperscript{2}Universitiy of Toronto, Toronto, Canada\\
{\tt\small m.elhoushi@ieee.org, zihao.chen@mail.utoronto.ca, farhan.shafiq@huawei.com,}\\ 
{\tt\small \{ye.henry.tian, li.joey922\}@gmail.com}}

\maketitle

\begin{abstract}
	The high computation, memory, and power budgets of inferring convolutional neural networks (CNNs) are major bottlenecks of model deployment to edge computing platforms, e.g., mobile devices and IoT. Moreover, training CNNs is time and energy-intensive even on high-grade servers. Convolution layers and fully connected layers, because of their intense use of multiplications, are the dominant contributor to this computation budget. 
	
	We propose to alleviate this problem by introducing two new operations: convolutional shifts and fully-connected shifts which replace multiplications with bitwise shift and sign flipping during both training and inference. During inference, both approaches require only 5 bits (or less) to represent the weights. This family of neural network architectures (that use convolutional shifts and fully connected shifts) is referred to as DeepShift models. We propose two methods to train DeepShift models: DeepShift-Q which trains regular weights constrained to powers of 2, and DeepShift-PS that trains the values of the shifts and sign flips directly. 
	
	Very close accuracy, and in some cases higher accuracy, to baselines are achieved. Converting pre-trained 32-bit floating-point baseline models of ResNet18, ResNet50, VGG16, and GoogleNet to DeepShift and training them for 15 to 30 epochs, resulted in Top-1/Top-5 accuracies higher than that of the original model. 
	Last but not least, we implemented the convolutional shifts and fully connected shift GPU kernels and showed a reduction in latency time of 25\% when inferring ResNet18 compared to unoptimized multiplication-based GPU kernels. The code can be found at \href{https://github.com/mostafaelhoushi/DeepShift}{https://github.com/mostafaelhoushi/DeepShift}. 
\end{abstract}

\section{Introduction}
Reducing the energy consumption, time latency, and memory requirements of deep neural networks for both inference and training are two requirements researchers are trying to achieve. For inference, deep neural networks are increasingly being targeted for mobile and IoT applications. Devices at the edge have a lower power and price budget as well as constrained memory size. Moreover, the amount of communication between different levels of memory in computing platforms also has a major role in the power requirements of a CNN. If communication between device and cloud becomes necessary (e.g. in case of model updates etc), model size could affect the connectivity costs. Therefore, for mobile / IoT inference applications, model optimization, size reduction, faster inference, and lower power consumption are key areas of research. 

For training, the energy required to train an average deep learning model on GPU servers is equivalent to using fossil fuel releasing around 78,000 pounds of carbon, which is more than half of a car's output during its lifetime \cite{Strubell2019EnergyAP}. According to a report by OpenAI \cite{AIandCompute}, the computation power required to train the largest neural network models doubles every 3.4 months, with an increase of 300,000$\times$ since 2012. Moreover, training on edge devices, e.g., in limited scenarios (including online learning, model adaptation, one/few-shot learning), is a growing field \cite{dhar2019ondevice,Tsukada_2020}. Therefore, for both cloud/server and edge / mobile applications, reducing energy and time consumption of training is required. 

Several approaches are being considered to address those needs and as such these efforts can be divided into a few categories. One approach is to design efficient models from the ground up resulting in novel network architectures that are more efficient to train and infer \cite{Iandola2017}. However, that proves to be a task requiring a lot of training resources to explore multiple variants of architectures to find the best fit. Another approach is to start with a big model initially and prune it \cite{frankle2018lottery, Elkerdawy_2019,li2020group, liu2019metapruning, 10.1007/978-3-030-58598-3_36}. \ifdefined\arxiv Pruning can be done by zeroing out individual parameters in model layers to reduce multiply-and-accumulate (MAC) operations (weight pruning) \cite{frankle2018lottery} or by removing whole blocks of parameters in convolutional layers (filter pruning) \cite{Elkerdawy_2019}. \fi

Although pruning increases the efficiency for inference, it does not increase the efficiency of training as it requires training larger models from the start. Another large category of techniques to accelerate inference and/or training is quantization, which this paper falls under. 

In such quantization techniques, the parameters of a model are converted from high-precision 32-bit floating-point representation to lower-precision smaller bit-width floating-point representations, such as 16-bits \cite{NVIDIAMixedPrecision} or 8-bits \cite{NIPS2019_8736, Jacob_2018_CVPR}, fixed-point representations \cite{pmlr-v37-gupta15}, or other representations with bit-widths that can range from 5-bits to 2-bits \cite{Lin2017BCNN,Zhou2016} and 1-bit \cite{NIPS2016_6573}. Some quantization techniques start with a full-precision pre-trained model (hence accelerating inference only)\cite{Lin2017BCNN} while others train from models from scratch and hence accelerating both inference and training \cite{Li2016}. Key attractions of these techniques are that they can be easily applied to various kinds of networks and they not only reduce the model size but also require less complex compute units on the underlying hardware. This results in a smaller model footprint, less working memory (and cache), faster computation on supporting platforms, and lower power consumption. Moreover, some quantization techniques replace multiplication with cheaper operations such as binary XNOR operations \cite{Lin2017BCNN, NIPS2016_6573,Liu_2018_ECCV} or bitwise shifts \cite{Zhou2017}. Binary XNOR techniques have high accuracy on small datasets such as MNIST or CIFAR10, but they suffer high degradation on complex datasets such as Imagenet \cite{Rastegari2016}. On the other hand, bit-wise shift techniques have proven to show a minimal drop in accuracy on Imagenet \cite{Zhou2017}.

This paper presents an approach to reduce computation and power requirements of CNNs by replacing regular multiplication-based convolution and linear operations, also known as a fully-connected layer or matrix multiplication, with bitwise-shift-based convolution and linear operations respectively. Applying bitwise shift operation on an element is mathematically equivalent to multiplying it by a power of 2. We compare our work with one similar approach that replaces multiplications with bitwise shift and sign flips: Incremental Network Quantization (INQ) \cite{Zhou2017}, and another approach that replaces multiplications with sign flips only: Accurate Binary Convolution (ABC). We propose two approaches: DeepShift-Q and DeepShift-PS. Both our DeepShift approaches are based on training the shift values (i.e. the powers of 2) and sign flips either indirectly or directly, while INQ is based on rounding 32-bit floating-point weights to powers of 2. To the best of our knowledge, this paper is the first to propose training the shift values directly, which is the main novelty of the paper. Moreover, INQ requires starting from a pre-trained model (hence can only accelerate inference) while both of our DeepShift proposed approaches can accelerate both inference and training.

This paper makes the following contributions:
\begin{itemize}
	\item We introduce two different methods to train DeepShift networks, neural networks with powers of 2 weights: each method trains the powers of 2 at run-time, and when computing the parameters' gradients at train-time (see Section \ref{sec:deepshift-networks}).
	\ifdefined\arxiv
	\item We conduct experiments for each training method, which show that it is possible to train DeepShift networks on CIFAR-10 and Imagenet, from scratch or from pre-trained full-precision weights and exceed or achieve nearly state-of-the-art results (see Section \ref{sec:benchmark-results}).
	\fi
	\item We develop bitwise-shift-based matrix multiplication and convolution GPU kernels that are able to run the DeepShift ResNet18 architecture with 25\% faster inference times than the baseline ResNet18 model with an unoptimized GPU kernel \ifdefined\arxiv(see Section \ref{subsec:gpu-implementation} )\fi.
\end{itemize}

\label{commented_sec:related-work}
\section{Related Work}
\ifdefined\arxiv
This section covers related works which fall into three categories. First, innovative new network architectures have been proposed that reduce the model sizes as well as computation requirements. \cite{Iandola2017} proposed SqueezeNet achieving 50x fewer parameters compared to AlexNet. \cite{Howard2017} proposed MobileNets, based on a streamlined architecture that uses depth-wise separable convolutions to build lightweight DNNs, followed by MobileNetV2 \cite{Sandler2018} introducing inverted residual structure where the input and output of the residual blocks are thin bottleneck layers opposite to traditional residual models. \cite{Chen2019DropAO} Proposed reducing spatial redundancy in CNN by replacing usual convolution operations with Octave Convolution, a spatially less-redundant variant.

Secondly, model quantization techniques have been an active research area and various approaches have been developed. \cite{Courbariaux2015} proposed Binary Connect constraining weights to only two values (-1 or 1). \cite{Rastegari2016} improve on Binary Connect, introducing Binary Weight Networks (BWNs), employing channel-wise scaling factors, while \cite{Li2016} introduced Ternary Weight Networks (TWNs). \cite{Zhu2017} extended TWN with Trained Ternary Quantization (TTQ), with non-uniform and trainable scaling factors. Other works like Binarized Neural Networks \cite{Courbariaux2016}, XNOR-Net \cite{Rastegari2016}, Bi-Real Net \cite{Liu2018} and ABCNet \cite{Lin2017BCNN} quantize both weights and activations to either -1 or 1. 
\fi

\cite{Zhou2017} proposed Incremental Network Quantization (INQ) to quantize pre-trained full-precision DNNs with weights constrained to zeros and powers of two. It replaces multiplications in convolutions with bitwise shifts and sign-flips but during inference only and not training. Hence, it cannot be used to reduce the energy consumption of on-device training (e.g., for transfer learning or zero-/one- shot training) nor on the cloud. INQ's approach involves iteratively partitioning the weights into two sets, one of which is quantized while the other is retrained to compensate for accuracy degradation. After each iteration, a bigger portion of the floating-point weights are converted to powers of 2, and the remaining portion is trained again. INQ introduces a new hyperparameter, the fraction of weights that are rounded to powers of 2 at each iteration. The authors of INQ listed the iterative steps they used for each CNN architecture, that was obtained or tuned empirically. Unlike INQ, our method does not introduce a new hyperparameter.

LogNN \cite{Miyashita2016ConvolutionalNN} is another method that converts weights as well as activations to powers of 2. However, it requires manual empirical tuning rather than training. The authors of LogNN did present an algorithm to train weights and activations in powers of 2, but it was only implemented on pre-trained models on CIFAR10 dataset. Also, LogNN does not convert the first layer, which may consume up to 20\% of the FLOPs in CNN architectures, to bitwise shifts, while our approach converts all layers. Furthermore, the authors did not explain how sign-flip of weights should be trained. The authors of LogNN also presented a hardware implementation in \cite{LogNet_7953288}. LogQuant \cite{Cai:2018:DLL:3291280.3291800} and ShiftCNN \cite{ShiftCNN} presented methods to convert the weights of a pre-trained model to powers of 2, such that no further training is needed and with a minimal decrease in accuracy. ShiftCNN replaces each multiplication with a group of 2 or 3 bitwise shifts, while our approach replaces each multiplication with a single bitwise shift. 


To the best of our knowledge, DeepShift is the first method to train a model from scratch using powers of 2.

\ifdefined\arxiv
\cite{Cai2017} proposed low-precision activations using Half-Wave Gaussian Quantization (HWGQ) while weights are binarized using BWN. \cite{Faraone2018} introduced Symmetric Quantization (SYQ) by pixel-wise scaling factors and fixed-point activation quantization. DoReFa-Net \cite{Zhou2016}, PACT \cite{Choi2018} and PACT-SAWB \cite{ChoiChunag2018} allows weights and activation to be variable configurable. \cite{Baskin2018} proposed uniform noise injection for non-uniform quantization (UNIQ) of both weights and activations. \cite{Lin2017} use multi-bit quantization. \cite{Zhang2018} proposed an adaptively learnable quantizer (LQ-Nets).

Thirdly pruning can be used to reduce model redundancy. Several pruning works have been proposed that employ various kinds of ranking mechanisms. \cite{Hanson1989} introduced hyperbolic and exponential biases for pruning in the late 80s. \cite{Hassibi1993} and \cite{LeCun1990} published some of the other earlier works on pruning. More recently, \cite{Han2016CoRR} proposed Deep Compression, a method that leverages pruning, weight sharing, and Huffman Coding for model compression. \cite{Lin2017} proposed Runtime Neural Pruning, a framework that prunes the deep neural network dynamically at the runtime. Pruning introduces another complexity to the mix since after pruning the computation units need to be able to handle sparse matrix arithmetic which adds some overhead. 

Finally, efficient kernel libraries \cite{GEMMLOWP} , \cite{IntelMKLDNN}, \cite{KeilCMSISNN}, \cite{QualcommSNPE}, \cite{NvidiaTensorRT} and custom hardware targeting reduced precesion have been developed to improve inference performance \cite{NVDLA}, \cite{Han2016ISCA}, \cite{Umuroglu2017}.
\fi

\section{DeepShift Networks} \label{sec:deepshift-networks}
\begin{figure*}
	\centering
	\begin{subfigure}[b]{.3\textwidth}
		\includegraphics[width=\textwidth]{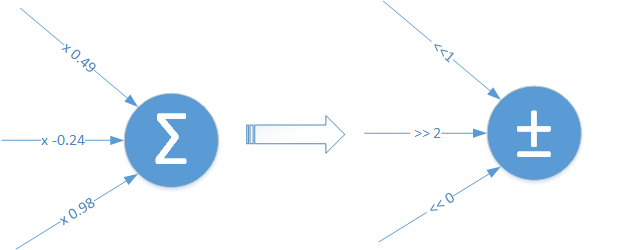}
		\caption{}\label{fig:shiftops:mac2sac}
	\end{subfigure}\qquad
	\begin{subfigure}[b]{.3\textwidth}
		\includegraphics[width=\textwidth]{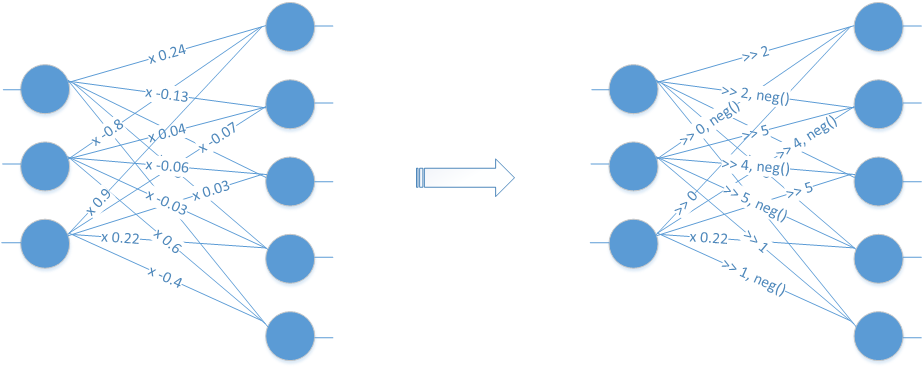}
		\caption{}\label{fig:shiftops:linear}
	\end{subfigure}\qquad
	\begin{subfigure}[b]{.3\textwidth}
		\includegraphics[width=\textwidth]{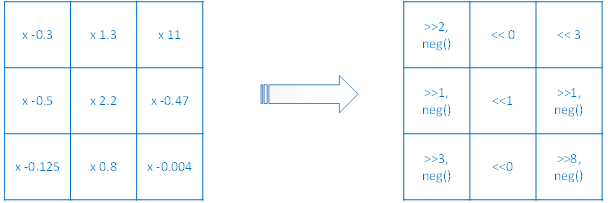}
		\caption{}\label{fig:shiftops:conv}
	\end{subfigure}
	\caption{(a) Multiply and Accumulate (MAC), which is the main operation in most neural networks, is replaced in DeepShift with bitwise shift and addition/subtraction. (b) Weights of original linear operator vs. proposed shift linear operator. (c) Weights of original convolution operator vs. proposed shift convolution operator}
	\label{fig:shiftops}
\end{figure*}

As shown in Figure \ref{fig:shiftops}, the main concept of this paper is to replace multiplication with bitwise shift and sign flip. If the underlying binary representation of an input number, $ x $ is in integer or fixed-point format, a bit-wise shift of $ \tilde{p} $ bits, where $ \tilde{p} $ is an integer, to the left (or right) is mathematically equivalent to multiplying by a positive (or negative power) of 2:
\begin{equation}
\begin{split}
2^{\tilde{p}}x = \begin{cases}
x << \tilde{p} \text{\; if } \tilde{p} > 0 \\
x >> \tilde{p} \text{\; if } \tilde{p} < 0 \\
x \text{\; if } \tilde{p} = 0
\end{cases} 
\end{split}
\text{such that} \text{\space} \tilde{p} \in \mathds{Z}
\end{equation}
For simplicity, we will use $ << \tilde{p} $ to implicitly denote a left shift if $ \tilde{p} $ is positive and right shift if $ \tilde{p} $ is negative.

Bitwise shift can only be equivalent to multiplying by a positive number, because $ 2^{\pm \tilde{p}} > 0 $ for any real value of $ \tilde{p} $. However, in neural networks, it is necessary for the training to have the equivalent of multiplying by negative numbers in its search space, especially in convolutional neural networks where filters with both positive and negative weights contribute to detecting edges. Therefore, we also need to use the sign flip (a.k.a negation) operation. We use a ternary sign operator, i.e., an operator that can either leave its operand unchanged, replace it with 0, or changes the sign of its operand. The sign flip operation can be represented mathematically as:
\begin{equation}
\begin{split}
\text{flip}(x, \tilde{s}) = 	\begin{cases}
-x \text{\; if } \tilde{s} = -1 \\
0 \text{\; if } \tilde{s} = 0 \\
x \text{\; if } \tilde{s} = +1  
\end{cases}
\end{split}
\text{such that} \text{\space} \tilde{s} \in \{-1, 0, +1\}
\end{equation}
Similar to bitwise shift, sign flip is a computationally cheap operation too as it involves returning the 2's complement of a number.

We introduce novel operators, LinearShift and ConvShift, that, in the forward pass, replace multiplication with bitwise shift and sign flip. Hence, the weight matrix, $ W $, whether it is used for linear operation (i.e., $ Y = WX + b $) or convolution operation (i.e., $ Y = W \circledast X + b $), will be replaced by elementwise product of the shift matrix $ \tilde{P} $ and sign matrix $ \tilde{S} $.
\begin{equation} \label{eqn:deepshift-fwd-pass}
W \to \text{flip}(2^{\tilde{P}}, \tilde{S})
\end{equation}

\begin{figure*}
	\centering
	\begin{subfigure}[b]{.4\textwidth}
		\includegraphics[width=\textwidth]{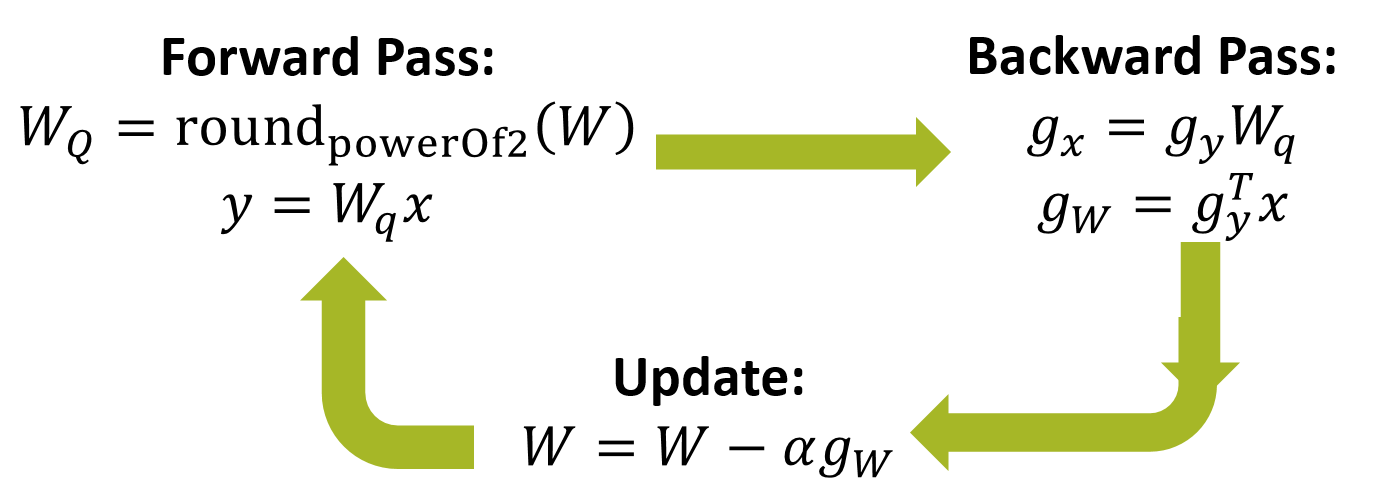}
		\caption{DeepShift-Q}\label{fig:training-method:deepshift-q}
	\end{subfigure}\qquad
	\begin{subfigure}[b]{.4\textwidth}
		\includegraphics[width=\textwidth]{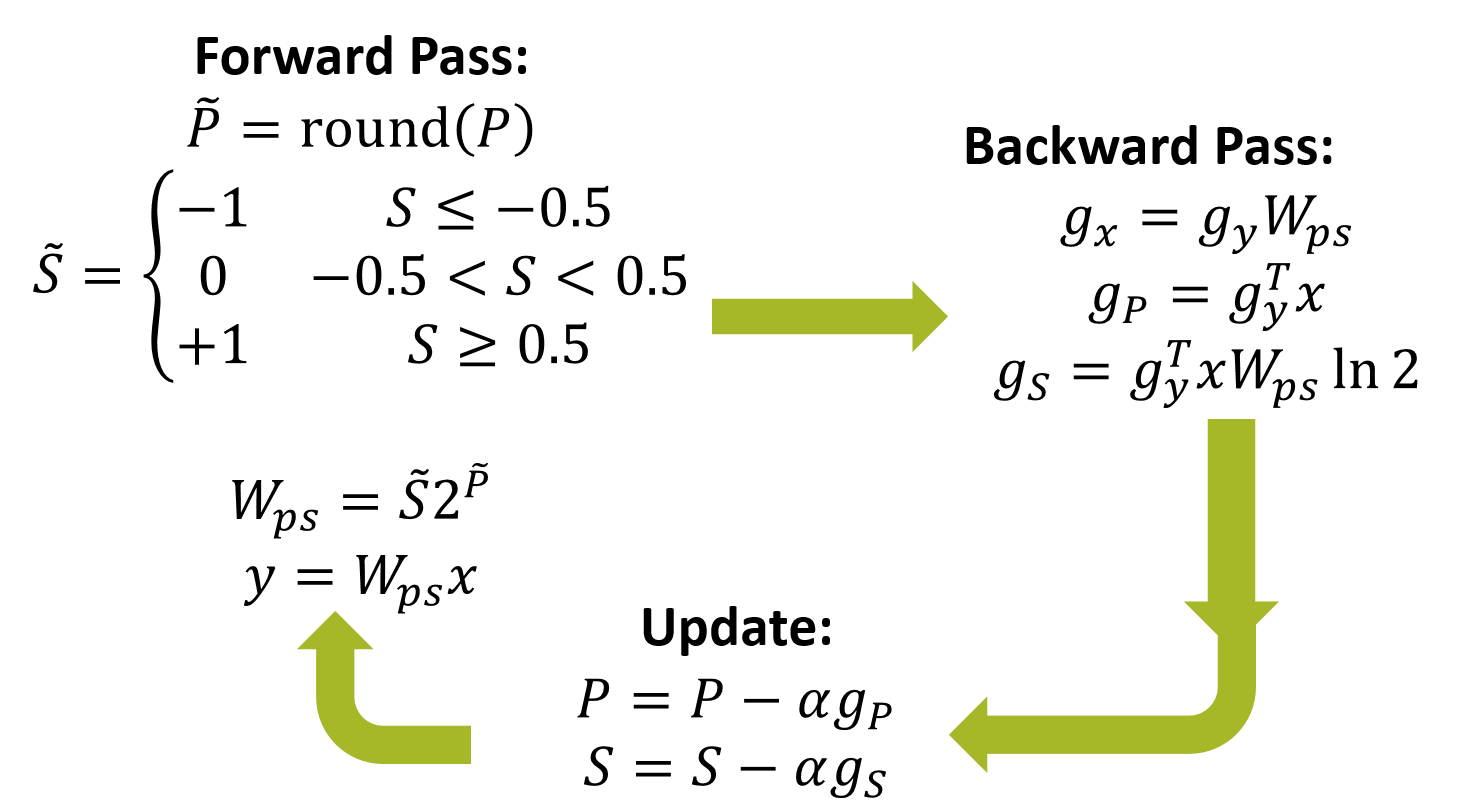}
		\caption{DeepShift-PS}\label{fig:training-method:deepshift-ps}
	\end{subfigure}
	\caption{We present 2 algorithms to train bitwise shift \& negation parameters of a model: (a) DeepShift-Q, (b) DeepShift-PS.} 
	\label{fig:training-method}
\end{figure*}

In the following sub-sections, we shall present two different methods to train the LinearShift and ConvShift operators of our DeepShift models: DeepShift-Q and DeepShift-PS. Both approaches use Equation \ref{eqn:deepshift-fwd-pass} for the forward pass but differ in the backward pass and differ in how $ \tilde{P} $ and $ \tilde{S} $ are derived. Both approaches are summarized in Figure \ref{fig:training-method}.

\subsection{DeepShift-Q}
In the DeepShift-Q approach, training is similar to the regular training approach of linear or convolution operations with weight $ W $, but in the forward pass and backward pass the weight matrix is quantized to $ \tilde{W}_{q} $ by rounding it to the nearest power of 2:
\begin{equation} \label{eqn:sign-equation}
\begin{split}
S &= \text{sign}(W) \\
\tilde{S} &= S
\end{split}
\end{equation} 

\begin{equation} \label{eqn:shift-equation}
\begin{split}
P &= \text{log}_2(\text{abs}(W)) \\
\tilde{P} &= \text{round}(P) \\
\end{split}
\end{equation} 

\begin{equation}
\begin{split}
\tilde{W}_{q} = \text{flip}(2^{\tilde{P}}, \tilde{S})
\end{split}
\end{equation} 

Hence, the forward pass for our LinearShift operator will be $ Y = \tilde{W}_{q}X = \text{flip}(2^{\tilde{P}}, \tilde{S})X $ and for our ConvShift operator will be $ Y = \tilde{W}_{q} \circledast X + b = \text{flip}(2^{\tilde{P}}, \tilde{S}) \circledast X + b $.

The gradients of the backward pass can be expressed as:
\begin{equation} \label{eqn:deepshift-q-bwd-pass}
\begin{split}
\frac{\partial C}{\partial X} & = \frac{\partial C}{\partial Y}\frac{\partial Y}{\partial X} = \frac{\partial C}{\partial Y}\tilde{W}_{q} \\
\frac{\partial C}{\partial W} & = \frac{\partial C}{\partial Y}\frac{\partial Y}{\partial \tilde{W}_{q}}\frac{\partial \tilde{W}_{q}}{\partial W} = \frac{\partial C}{\partial Y}X\frac{\partial \tilde{W}_{q}}{\partial W} \\
\frac{\partial C}{\partial b} & = \frac{\partial C}{\partial Y}
\end{split}
\end{equation}
where $ \frac{\partial C}{\partial Y} $ is the output gradient to the operator (derivative of the model loss (a.k.a cost function), $ C $, with respect to the operator output), $ \frac{\partial C}{\partial X} $ is the input gradient to the operator (derivative of the model loss with respect to the operator input)\footnote{We have simplified the expression of $ \frac{\partial C}{\partial X} $ in Equation \ref{eqn:deepshift-q-bwd-pass}. To be exact, it is equivalent to $ \frac{\partial C}{\partial Y}\tilde{W}_{q}^T $ for LinearShift and $ \tilde{W}_{q}^{rot180^{\circ}} \circledast \frac{\partial C}{\partial Y} $ for ConvShift.}, and $ \frac{\partial C}{\partial W} $ is the derivative of the model loss with respect to the operator weights \footnote{For LinearShift, $ \frac{\partial C}{\partial W} = X^T\frac{\partial C}{\partial Y} $ and for ConvShift $ \frac{\partial C}{\partial W} = X \circledast \frac{\partial C}{\partial Y} $}.

Since $ W_q $ is a rounded value of $ W $, the differentiation $ \frac{\partial \tilde{W}_{q}}{\partial W} $ is zero at all points except at the points where $ W $ is a power of 2. This will result in $ W $ not being updated during backpropagation, and hence no learning or updating of $ W $. We solve this issue by using a straight-through estimator (STE) \cite{yin2018understanding}:
\begin{equation}
\frac{\partial \tilde{W}_{q}}{\partial W} \approxeq 1
\end{equation} 
Hence, $ \frac{\partial C}{\partial W} $ is set to:
\begin{equation}
\frac{\partial C}{\partial W} \approxeq X\frac{\partial Y}{\partial \tilde{W}_{q}}
\end{equation} 

Figure \ref{fig:training-method:deepshift-q} summarizes the forward and backward passes of DeepShift-Q.

\subsection{DeepShift-PS}
DeepShift-PS on the other hand directly uses the shift, $ P $, and sign flip, $ S $, as the trainable parameters:
\begin{equation}
\begin{split}
\tilde{P} &= \text{round}(P) \\
\tilde{S} &= \text{sign}(\text{round}(S)) \\
\tilde{W}_{ps} &= \text{flip}(2^{\tilde{P}}, \tilde{S})
\end{split}
\end{equation}

Note the definition of sign flip operation, $ \tilde{S} $, implies that it is a ternary value rather than a binary parameter, as each element in the matrix can take one of 3 values:
\begin{equation}
\tilde{s} = \begin{cases}
-1 \text{\; if } s \leq -0.5 \\
0 \text{\; if } -0.5 < s < 0.5 \\
+1 \text{\; if } s \geq 0.5  
\end{cases}
\end{equation}
where $ s $ is an element of matrix $ S $ and $ \tilde{s} $ is an element of $ \tilde{S} $.

The gradients of the backward pass can be expressed as:
\begin{equation}
\begin{split}
\frac{\partial C}{\partial X} &= \frac{\partial C}{\partial Y}\frac{\partial Y}{\partial X} \\
&= \frac{\partial C}{\partial Y}W_{ps}^T \\
\frac{\partial C}{\partial P} &= \frac{\partial C}{\partial Y}\frac{\partial Y}{\partial \tilde{W}_{ps}}\frac{\partial \tilde{W}_{ps}}{\partial \tilde{P}}\frac{\partial{\tilde P}}{\partial P} \\
\frac{\partial C}{\partial S} & = \frac{\partial C}{\partial Y}\frac{\partial Y}{\partial \tilde{W}_{ps}}\frac{\partial \tilde{W}_{ps}}{\partial \tilde{S}}\frac{\partial \tilde{S}}{\partial S} \\
\frac{\partial C}{\partial b} & = \frac{\partial L}{\partial Y}
\end{split}
\end{equation}

We use the straight through estimators to express the derivatives of the round function and the sign function: $ \partial \text{round}(x) \approxeq \partial x \text{ , } \partial \text{sign}(x) \approxeq \partial x $. Hence, $ \frac{\partial \text{round}(x)}{\partial x} \approxeq 1 \text{ , } \frac{\partial \text{sign}(x)}{\partial x} \approxeq 1 $. However, for the flip function we use a modified straight through estimator: $\frac{\partial \text{flip}(x,s)}{\partial x} \approxeq \text{flip}(x,s) \text{ , } \frac{\partial \text{flip}(x,s)}{\partial s} \approxeq 1 $.

Based on that, we set $ \frac{\partial \tilde{P}}{\partial P} \approxeq 1 $ and $ \frac{\partial \tilde{S}}{\partial S} \approxeq 1 $. Hence, $ \frac{\partial \tilde{W}_{ps}}{\partial \tilde{S}} $ is evaluated as:
\begin{equation}
\frac{\partial \tilde{W}_{ps}}{\partial \tilde{S}} = \frac{\partial \text{flip}(2^{\tilde{P}}, \tilde{S})}{\partial \tilde{S}} \approxeq 1 \\
\end{equation}

and, $ \frac{\partial \tilde{W}_{ps}}{\partial \tilde{P}} $ is evaluated as:
\begin{equation}
\begin{split}
\frac{\partial \tilde{W}_{ps}}{\partial \tilde{P}} &= \frac{\partial \text{flip}(2^{\tilde{P}}, \tilde{S})}{\partial \tilde{P}} = \frac{\partial \text{flip}(2^{\tilde{P}}, \tilde{S})}{\partial (2^{\tilde{P}})} \frac{\partial (2^{\tilde{P}})}{\partial \tilde{P}}  \\
& \approxeq \text{flip}(2^{\tilde{P}}, \tilde{S}) 2^{\tilde{P}} \text{ln}2 = \tilde{W}_{ps} \text{ln}2 \\
\end{split}
\end{equation}

Hence, the gradients of the weights with respect to loss are set to:
\begin{equation}
\begin{split}
\frac{\partial C}{\partial P} &\approxeq \frac{\partial C}{\partial Y}\frac{\partial Y}{\partial \tilde{W}_{ps}}\tilde{W_{ps}} \text{ln}2 \\
\frac{\partial C}{\partial S} &\approxeq \frac{\partial C}{\partial Y}\frac{\partial Y}{\partial \tilde{W}_{ps}} 
\end{split}
\end{equation}

Figure \ref{fig:training-method:deepshift-ps} summarizes the forward and backward passes of DeepShift-PS.
\section{Implementation}
We implement the forward and backward passes of our two custom ops, LinearShift and ConvShift, using PyTorch. Similar to \cite{NIPS2016_6573}, we emulate the precision of an actual bitwise shift hardware implementation by rounding the input and bias of both ops to 32-bit fixed-point format precision (16-bit for integer part and 16-bit for fraction part) before applying the forward pass.

To initialize the weights, we use Kaiming initialization \cite{he2015delving} for $ W_q $ in DeepShift-Q, and uniform random distribution to initialize $ P $ and $ S $ in DeepShift-PS.

In DeepShift-PS, L2 normalization has been slightly modified to occur on $ W_{PS} $ rather than on $ P $ on $ S $, i.e. the regularization term added is $ \sum{W_{PS}^2} = \sum{{2^PS}^2} $ rather than $ \sum{P^2} + \sum{S^2} $. This modification of L2 normalization was necessary to avoid gradient descent from pushing shift parameters, $ P $, from values whose squares are large, like -14 or -8, but are actually equivalent to multiplication with small weights, $ 2^{-8} $ and $ 2^{-14} $, to values whose squares are small like 0 or 1, but are actually equivalent to multiplying with relatively large weight values, 1.0 or 0.5, respectively.

Since activations are represented as 32-bit fixed point activations, the theoretical range of shift values that we need to support are -31 to +31, and this requires $ \text{log}_2(31 - (-31) + 1) = \text{log}_2(63) \approxeq 6 $ bits to represent shift. However, we noticed that shift values that result from training are rarely positive (i.e., most of the time $ \text{abs}(W_Q) < 1 $, hence $ P < 0 $ as $ P = \text{log}_2(\text{abs}(\text{round}(W_Q))) $ \footnote{This expression holds for both $ W_Q $ and $ W_{PS} $}). Further testing that we performed showed that high accuracy can be retained if we constrain shift values to be between 0 and -14 or 0 and -15. Therefore, we only need 4 bits to represent shift. Adding the bit required for the sign flip operation, we need 5 bits in total. During training, shift values that fell outside the supported shift ranges were clipped. In some of the results that we will explain in the upcoming sections, we have also experimented with fewer bits to represent weight, i.e., constraint shift values to shorter ranges.


\section{Benchmark Results} \label{sec:benchmark-results}
We have tested the training and inference results on 2 datasets: CIFAR10 \cite{cifar10_report}, and Imagenet \cite{imagenet_cvpr09}. For each dataset, we have tested a group of architectures. We have explored various evaluations for the models:
\begin{enumerate}
	\item \textbf{Original Version}: evaluating the original architecture with standard convolution and linear operators, \label{enum:eval_original}
	\item \textbf{DeepShift Versions}: 
	\begin{enumerate}
		\item \textbf{Train from Scratch}: start with randomly initialized weights, convert all the convolution and linear operators to their shift counterparts, and train from scratch using either training method Deepshift-Q or DeepShift-PS, \label{enum:train_shift_scratch}
		\item \textbf{Train from Pre-trained Baseline}: start with baseline model that is pretrained with 32-bits floating points weights, convert all the convolution and linear operators to their shift counterparts, convert the weights (using Equations \ref{eqn:sign-equation} and \ref{eqn:shift-equation}) and train using either training method Deepshift-Q or DeepShift-PS. 
	\end{enumerate}
\end{enumerate}

We have noticed - in terms of validation accuracy results - that the best optimizer for DeepShift-PS is Rectified Adam (RAdam) \cite{liu2019radam} while the best optimizer for DeepShift-Q and regular 32-bit floating-point training is stochastic gradient descent (SGD). \ifdefined\arxiv On the other hand, most methods in literature that explore alternatives to multiplication or quantizing weight representations use SGD only. We will show in one of our tests in CIFAR10 that using RAdam optimizer does not add an unfair advantage over other methods. \fi

\ifdefined\arxiv
\label{commented_sec:mnist}
\subsection{MNIST Data Set}
To verify the performance of DeepShift on a small dataset like MNIST, two simple models were trained and tested:
\begin{itemize}
	\item \textbf{Simple FC}: a simple fully-connected model consisting of 3 linear layers with feature output sizes 512, 512, and 10 respectively. Dropout layers with a probability of 0.2 were inserted in between the layers. All intermediate layers had a ReLu activation following it. 
	\item \textbf{Simple CNN}: a model consisting of 2 convolutional layers and 2 linear layers. The 2 convolutional layers had output channels sizes 20 and 50 respectively, and both had kernel sizes of 5x5 and strides of 1. Max pooling layers of window size 2x2 followed by ReLu activation were inserted after each convolution layer. The linear layers had output feature sizes of 500 and 10 respectively. 
\end{itemize}

A learning rate of 0.01, a momentum of 0.0, as well as a batch size of 64, was used to train. The training plot is shown in Figure \ref{fig:mnist}. The accuracy on the validation set is shown in Table \ref{mnist-results}. 

\begin{figure*}
	\centering
	\begin{subfigure}[b]{.4\textwidth}
		\includegraphics[width=\textwidth]{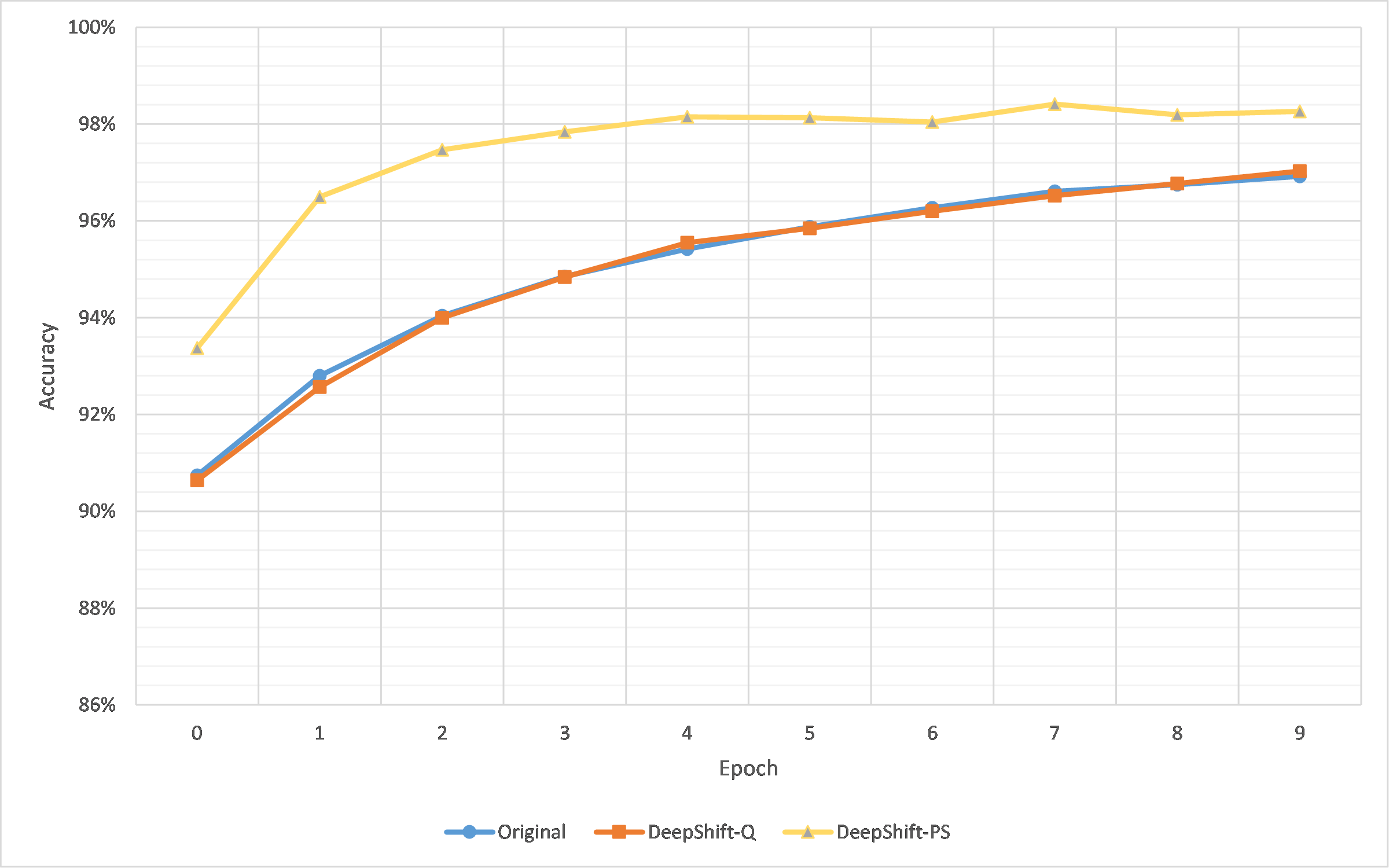}
		\caption{Simple FC}\label{fig:mnist:linear}
	\end{subfigure}\qquad
	\begin{subfigure}[b]{.4\textwidth}
		\includegraphics[width=\textwidth]{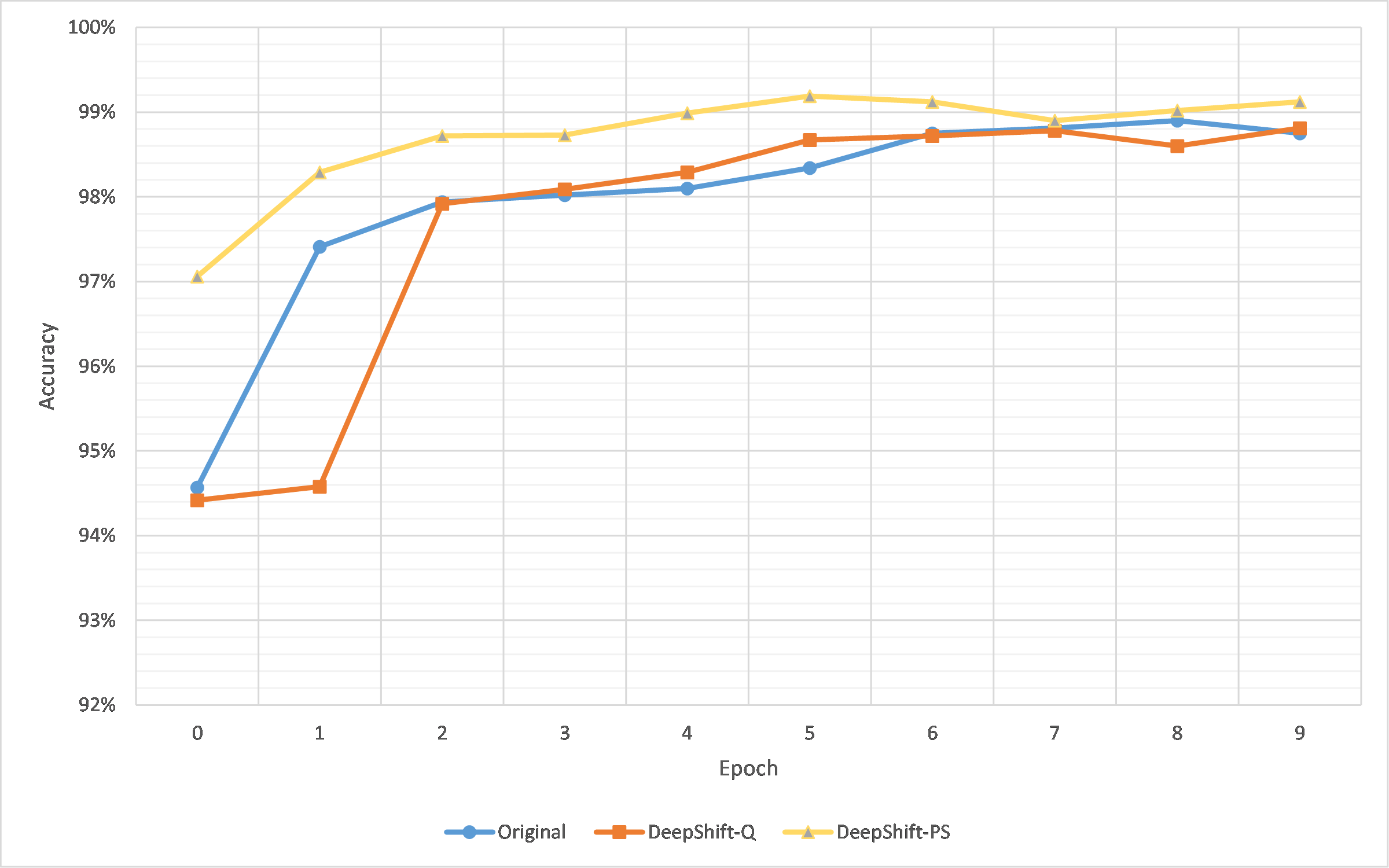}
		\caption{Simple CNN}\label{fig:mnist:conv}
	\end{subfigure}
	\caption{MNIST training from scratch}
	\label{fig:mnist}
\end{figure*}

\begin{table}
	\caption{MNIST Accuracy Results}
	\label{mnist-results}
	\centering
	\begin{tabularx}{\columnwidth}{l*{2}{>{\centering\arraybackslash}X}}
		\toprule
		Method & \shortstack[l]{Train \\ from Scratch} & \shortstack[l]{Train \\ from Pre-trained} \\ 
		\cmidrule(r){1-3}
		Simple FC & & \\
		\midrule
		Original & 96.92\%  & -      \\
		DeepShift-Q & 97.03\%  & 94.91\%    \\
		DeepShift-PS & 98.26\%  & 98.26\%   \\
		\midrule
		Simple CNN & & \\
		\midrule
		Original & 98.75\%  & -      \\
		DeepShift-Q & 98.81\%  & 99.15\%    \\
		DeepShift-PS & 99.12\%  & 99.16\%   \\
		\bottomrule
	\end{tabularx}
\end{table}
\fi

\subsection{CIFAR10 Data Set}
ResNet18 \cite{He_2016_CVPR} and MobileNetv2 \cite{Sandler2018} models were trained from scratch on CIFAR10 using both DeepShift methods. The models were trained with a momentum of 0.9 and weight decay of \num{1e-4}. The loss criterion used was categorical cross-entropy. The learning rate used to train was 0.1 that decays by a factor of 0.1 at epochs 80, 120, and 160, the number of epochs for training was 200, and the batch size was 128. 

For ResNet18, we have added an additional test of training the baseline (multiplication-based) 32-bit floating-point weights using RAdam, to compare it with DeepShift-PS that uses RAdam. This test was made to ensure that the comparisons are fair and that the use of RAdam as an optimizer is not giving an unfair advantage to DeepShift-PS. The best accuracy reached by the baseline ResNet18 model with SGD was 94.45\% and with RAdam was 94.06\%.

For training starting with a pre-trained model, the learning rate was $ 1 \times 10^{-3} $, decaying with a rate of 0.1 every 5 epochs, and the number of epochs was 15.

In order to compare with other multiplication-less models, we also trained VGG19-Small \cite{simonyan2015deep} from scratch to compare with AdderNet \cite{AdderNet}, that replaces multiplications with additions, and ShiftAddNet \cite{ShiftAddNet} that combined our DeepShift-PS approach with AdderNet. In Table \ref{cifar10-results} we mark ShiftAddNet with an asterisk to indicate that the accuracies shown are based on the results from the authors' paper rather than being reproduced by us. In order to make a fair comparison, the results shown are based on using 32-bit fixed-point activations.

\ifdefined\arxiv The training plots are shown in \ref{fig:cifar10}. \fi The numerical results for evaluating the validation set for various scenarios are shown in Table \ref{cifar10-results}. The results show that on CIFAR10, training with bit-wise shifts is capable is obtaining results almost the same as, and in the case of MobileNetv2 with DeepShift-Q, more than the multiplication-based baseline.

\begin{table}
	\caption{CIFAR10 Accuracy Results}
	\label{cifar10-results}
	\centering
	\begin{tabularx}{\columnwidth}{ll*{2}{>{\centering\arraybackslash}X}}
		\toprule
		Model & Method & Train from Scratch & Train from Pre-trained \\ 
		\midrule 
		ResNet18 & Original & 94.45\%  & -      \\
		& DeepShift-Q & 94.42\%  & 94.25\%    \\
		& DeepShift-PS & 93.20\%  & 94.12\%   \\
		\midrule
		MobileNet & Original & 93.57\%  & -      \\
		& DeepShift-Q & 93.63\%  & 93.04\%    \\
		& DeepShift-PS & 92.64\%  & 92.78\%   \\
		\midrule
		VGG19-Small & Original & 92\%  & -      \\
		& DeepShift-PS & 91.57\%  & N/A   \\
		& AdderNet \cite{AdderNet} & 93.02\% & N/A \\
		& ShiftAddNet* \cite{ShiftAddNet} & 90\% & N/A \\
		\bottomrule
	\end{tabularx}
\end{table}

\ifdefined\arxiv
\begin{figure*}
	\centering
	\begin{subfigure}[b]{.4\textwidth}
		\includegraphics[width=\textwidth]{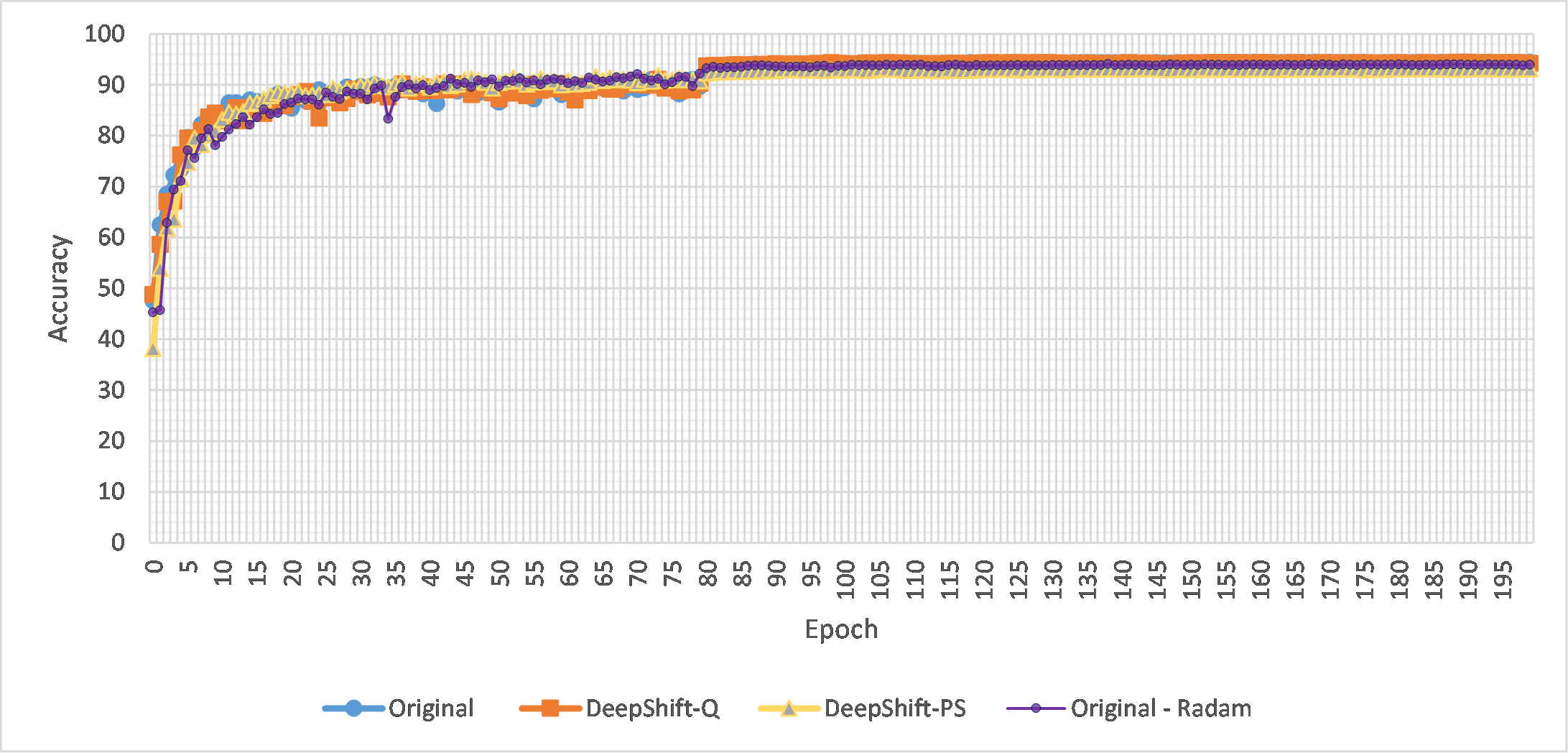}
		\caption{ResNet18}\label{fig:cifar10:resnet18}
	\end{subfigure}\qquad
	\begin{subfigure}[b]{.4\textwidth}
		\includegraphics[width=\textwidth]{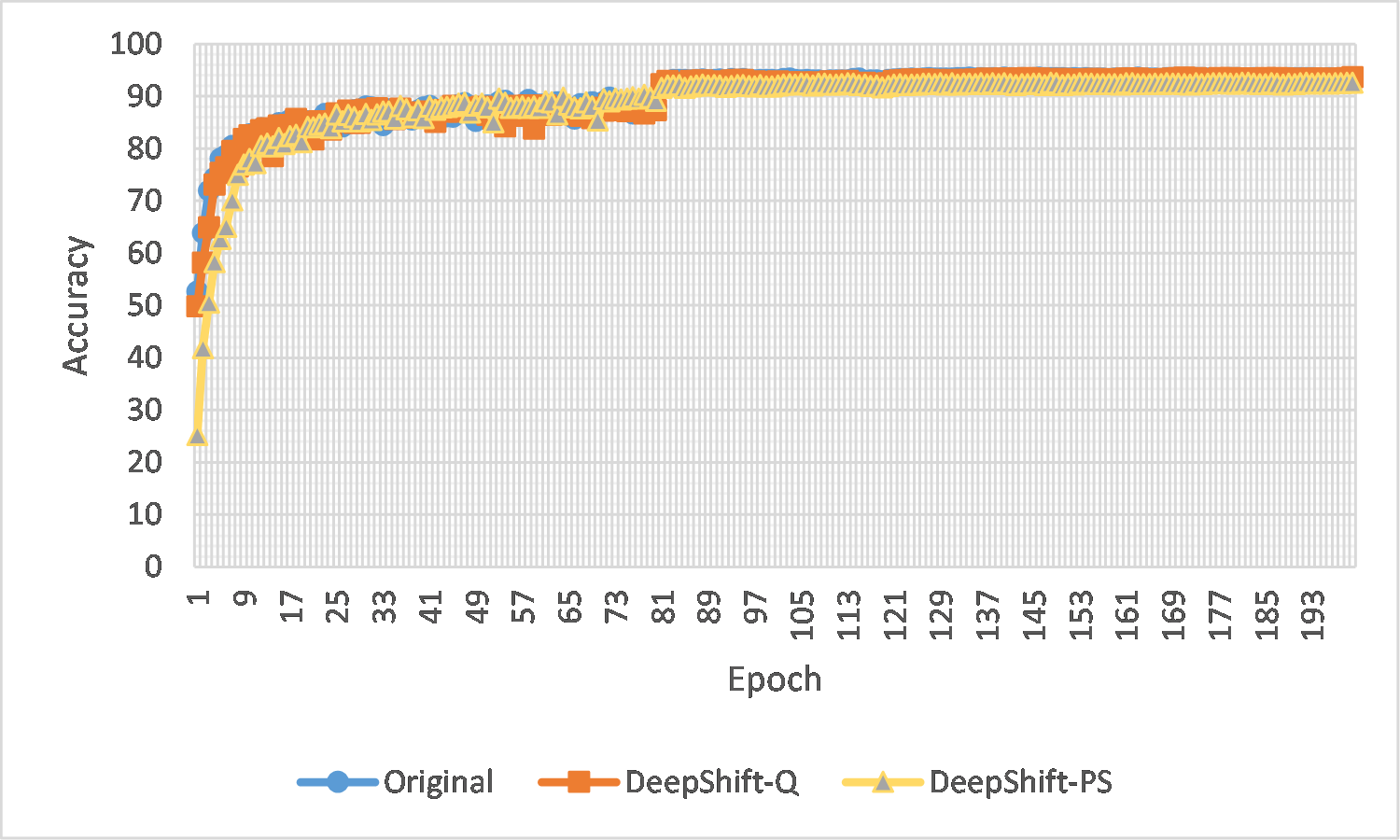}
		\caption{MobileNet}\label{fig:cifar10:mobielenet}
	\end{subfigure}
	\caption{CIFAR10 training from scratch}
	\label{fig:cifar10}
\end{figure*}
\fi

\subsubsection{Lower Bit Widths} \label{sec:lower-bit-wdiths}
Table \ref{cifar10-results-lower-bitwidths} shows the results for using fewer bits to represent weights or activations. For 16-bit activations, we used a fixed-point representation with 3-bits to represent the integer part, and 13-bits to represent the fraction part. Similarly for 8-bit activations, we used 3-bits to represent the integer part, and 13-bits to represent the fraction part. For training 8-bit or 16-bit activations from scratch, we changed the initial learning rate to 0.001.

We can see that for 3-bits and 2-bits weights, training from scratch obtained higher accuracy than training from a pre-trained baseline. This is because, converting the pre-trained weights to powers of 2 constrained to a short range, caused many weight values to be clipped to the same value. Hence the distribution of weights of a model that is randomly initialized, i.e., training from scratch, is better than weights of a pre-trained model clipped to small values \cite{Mishkin2015AllYN}. \ifdefined\arxiv Moreover, the results show that DeepShift-PS outperforms DeepShift-Q when training from scratch for smaller weight bit-widths. \fi

On the other hand, DeepShift-Q outperforms DeepShift-PS when training from scratch using 16-bit or 8-bit activations, while both are close to baseline accuracies when trained from pre-trained weights.

\begin{table}
	\caption{ResNet18 CIFAR10 Accuracy Results for Lower Bitwidths}
	\label{cifar10-results-lower-bitwidths}
	\centering
	\begin{tabularx}{\columnwidth}{lll*{2}{>{\centering\arraybackslash}X}}
		\toprule
		Method & W & A & Train from Scratch & Train from Pre-trained \\ 
		\midrule
		Original & 32 & 32 & 94.45\%  & -      \\
		\midrule
		DeepShift-PS & 5 & 32 & 93.20\% & 94.12\%    \\
		             & 5 & 16 & 85.54\% & 94.15\% \\
		             & 5 & 8  & 86.28\% & 93.96\% \\
		             & 4 & 32 & 94.12\% & 94.13\%   \\
		             & 3 & 32 & 92.85\% & 91.16\%    \\
		             & 2 & 32 & 92.80\% & 90.68\%   \\
        \midrule
        DeepShift-Q & 5 & 32 & 94.60\% & 94.43\% \\	
		            & 5 & 16 & 94.06\% & 93.70\% \\
            		& 5 & 8  & 94.14\% & 93.65\% \\
		            & 4 & 32 & 94.11\% & 94.38\% \\
		            & 3 & 32 & 88.28\% & 92.04\% \\
		            & 2 & 32 & 69.97\% & 89.96\% \\
		\bottomrule
	\end{tabularx}
\end{table}


\subsection{Imagenet Data Set}
We categorize methods in literature that accelerate neural networks by replacing multiplications with cheaper operations or by quantization, into methods: 
\begin{itemize}
	\item that train from pre-trained baseline: INQ \cite{Zhou2017}, ABC \cite{Lin2017BCNN}, and
	\item that train from scratch, BWN \cite{Rastegari2016}, TWN \cite{Li2016}.
\end{itemize}
In each of the following two sub-sections, we compare DeepShift to each category. Results for various CNN architectures for both categories are shown in Tables \ref{imagenet-results-from-pretrained} and \ref{imagenet-results-from-scratch}. Since different papers report different accuracies for full-precision baselines, we either attempt to reproduce the results of a method or report the difference between the accuracy of the method and its corresponding baseline (method names marked with asterisks).

\subsubsection{Training from Pre-Trained Baseline}
Table \ref{imagenet-results-from-pretrained} compares DeepShift trained from a pre-trained baseline to INQ and ABC methods. Column ``W'' shows the number of bits to represent weights in each method. The DeepShift models were trained with momentum of 0.9, weight decay of \num{1e-4}, the loss criterion categorical cross-entropy, the initial learning rate was 0.001 that decays by a factor of 0.1 every 5 epochs. 

INQ \cite{Zhou2017} (Incremental Network Quantization) replaces multiplications in convolutions with bitwise shifts and sign-flips, but during inference only and not training. It also introduces a hyperparameter, a list of portions per iteration to round the weights to power of 2. This list of portions is obtained in an empirical manner for each architecture. DeepShift on the other hand does not introduce any hyperparameter. INQ was originally developed using Caffe, and the pre-trained baselines reported were significantly lower than what we have using PyTorch. Hence, to start with the same baseline, we used the PyTorch implementation of INQ \cite{INQPyTorch} to report the accuracies in Table \ref{imagenet-results-from-pretrained}. In terms of accuracies, both INQ and DeepShift exceed or almost meet the baseline accuracies of most architectures. For ResNet18, we also compare using 4-bits to represent weights.

ABC \cite{Lin2017BCNN} (Accurate Binary Convolution) neural networks replace multiplication with XNOR (i.e., sign flip) operation. Comparing with the same number of bits, 5, for weight representation, DeepShift obtains higher accuracy than ABC.



\begin{table*}[!t]
	\caption{Imagenet Accuracy Results: Training from Pre-trained Baseline. ``W'' refers to number of bits to represent weights. \ifdefined\arxiv``Top-N'' accuracy means that the correct class gets to be in the Top-N probabilities for it to count as ``correct''. \fi}
	\label{imagenet-results-from-pretrained}
	\centering
	\begin{tabularx}{\textwidth}{ll*{5}{>{\centering\arraybackslash}X}}
		\toprule
		Model & Method & W & \multicolumn{2}{c}{Top-1} & \multicolumn{2}{c}{Top-5} \\ 
		\cmidrule(r){4-7}
		& & & Accuracy & Delta & Accuracy & Delta \\
		\midrule
		\bottomrule
		ResNet18 & Original & 32 & 69.76\%  & - & 89.08\% & - \\
		& DeepShift-Q [Ours] & 5 & 69.56\% & -0.20\% & 89.17\% & +0.09\%  \\
		& DeepShift-PS [Ours] & 5 & 69.27\% & -0.49\% & 89.00\% & -0.08\%  \\
		& INQ \cite{Zhou2017} & 5 & 69.36\% & -0.40\% & 89.09\% & +0.01\% \\
		& ABC* \cite{Lin2017BCNN} & 5 & 68.30\% & -1.00\% & 87.90\% & -1.30\%  \\
		\midrule
		& DeepShift-Q [Ours] & 4 & 69.56\% & -0.20\% & 89.14\% & +0.06\%  \\
		& DeepShift-PS [Ours] & 4 & 69.02\% & -0.74\% & 88.73\% & -0.35\%  \\
		& INQ* \cite{Zhou2017} & 4 & 68.88\% & +0.63\% & 89.01\% & +0.32\% \\
		\midrule
		\bottomrule
		ResNet50 & Original & 32 & 76.13\% & - & 92.86\% & - \\
		& DeepShift-Q [Ours] & 5 & 76.33\% & +0.22\% & 93.05\% & +0.19\%  \\
		& DeepShift-PS [Ours] & 5 & 75.93\% & -0.20\% & 92.90\% & +0.04\%  \\
		& INQ \cite{Zhou2017} & 5 & 75.01\% & -1.12\% & 92.45\% & -0.41\% \\
		\midrule
		\bottomrule
		GoogleNet & Original & 32 & 69.78\%  & - & 89.53\% & - \\
		& DeepShift-Q [Ours] & 5 & 70.73\% & +0.95\% & 90.17\% & +0.64\%    \\
		& DeepShift-PS [Ours] & 5 & 69.87\% & +0.09\% & 89.62\% & +0.09\%   \\
		& INQ \cite{Zhou2017} & 5 & 71.25\% & +1.47\% & 90.33\% & +0.70\% \\
		\ifdefined\arxiv
		& LogQuant* \cite{Cai:2018:DLL:3291280.3291800} & 6 & 69.14\% & -0.36\% & 88.86\% & -0.28\% \\
		\fi
		\midrule
		\bottomrule
		VGG16 & Original & 32 & 71.59\% & - & 90.38\% & - \\
		& DeepShift-Q [Ours] & 5 & 71.56\% & -0.03\% & 90.48\% & +0.10\%  \\
		& DeepShift-PS [Ours] & 5 & 71.39\% & -0.20\% & 90.33\% & -0.05\%  \\
		& INQ \cite{Zhou2017} & 5 & 71.32\% & -0.27\% & 90.29\% & -0.09\% \\
		\ifdefined\arxiv
		& LogQuant* \cite{Cai:2018:DLL:3291280.3291800} & 6 & N/A & N/A & 89.62\% & -0.31\% \\ 
		\fi
		\midrule
		\bottomrule
		AlexNet & Original & 32 & 56.52\% & - & 79.07\% & - \\
		& DeepShift-Q [Ours] & 5 & 55.81\% & -0.71\% & 78.79\% & -0.28\%  \\
		& DeepShift-PS [Ours] & 5 & 55.90\% & -0.62\% & 78.73\% & -0.34\%  \\
		& INQ \cite{Zhou2017} & 5 & 56.13\% & -0.39\% & 78.83\% & -0.24\% \\
		\ifdefined\arxiv
		\fi
		\midrule
		\bottomrule
	\end{tabularx}
\end{table*}

\subsubsection{Training from Scratch}

\begin{figure*}
	\centering
	\ifdefined\arxiv
	\begin{subfigure}[b]{.4\textwidth}
		\includegraphics[width=\textwidth]{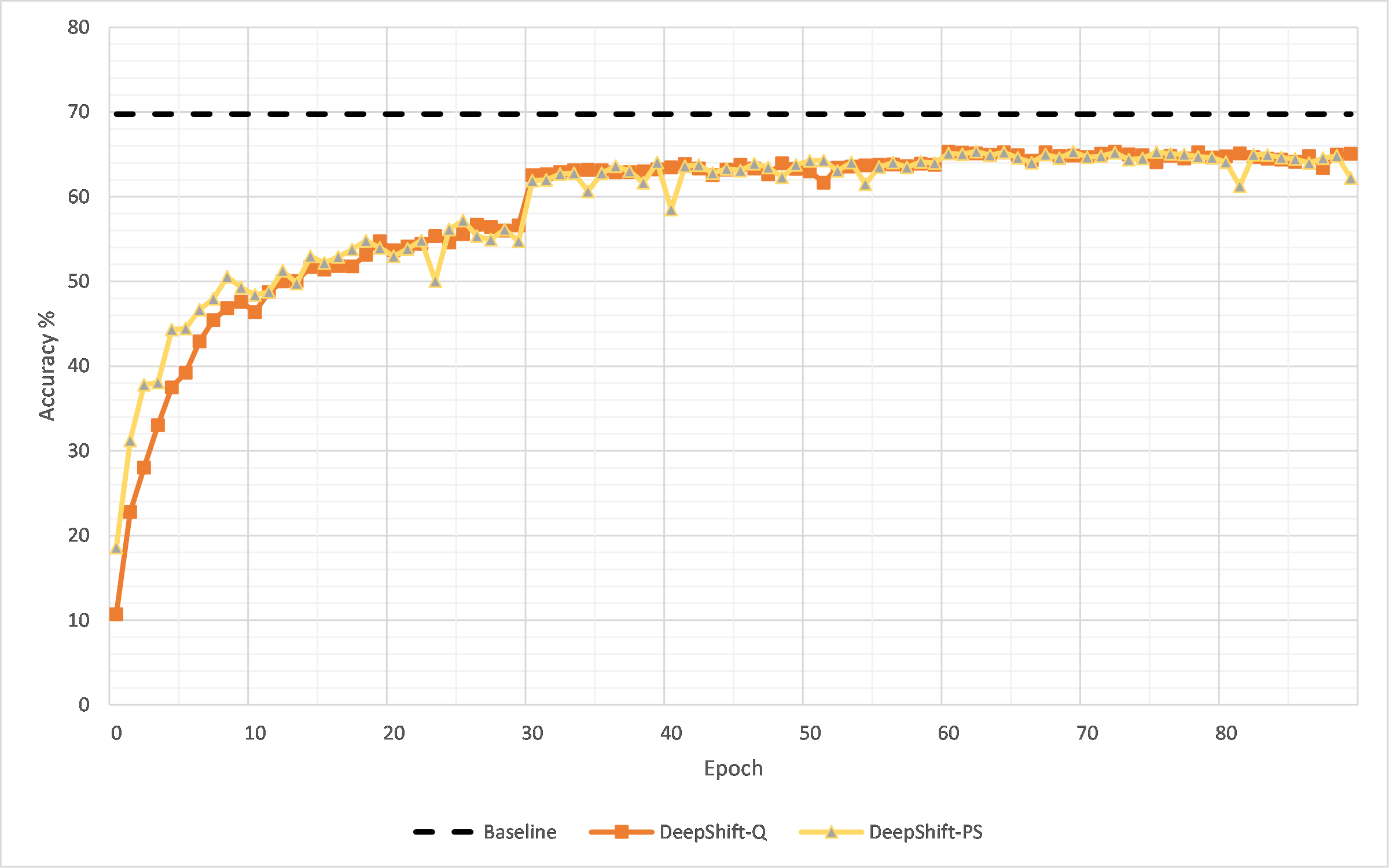}
		\caption{ResNet18 Top-1 Accuracy}\label{fig:imagenet:resnet18-top1}
	\end{subfigure}\qquad
	\begin{subfigure}[b]{.4\textwidth}
		\includegraphics[width=\textwidth]{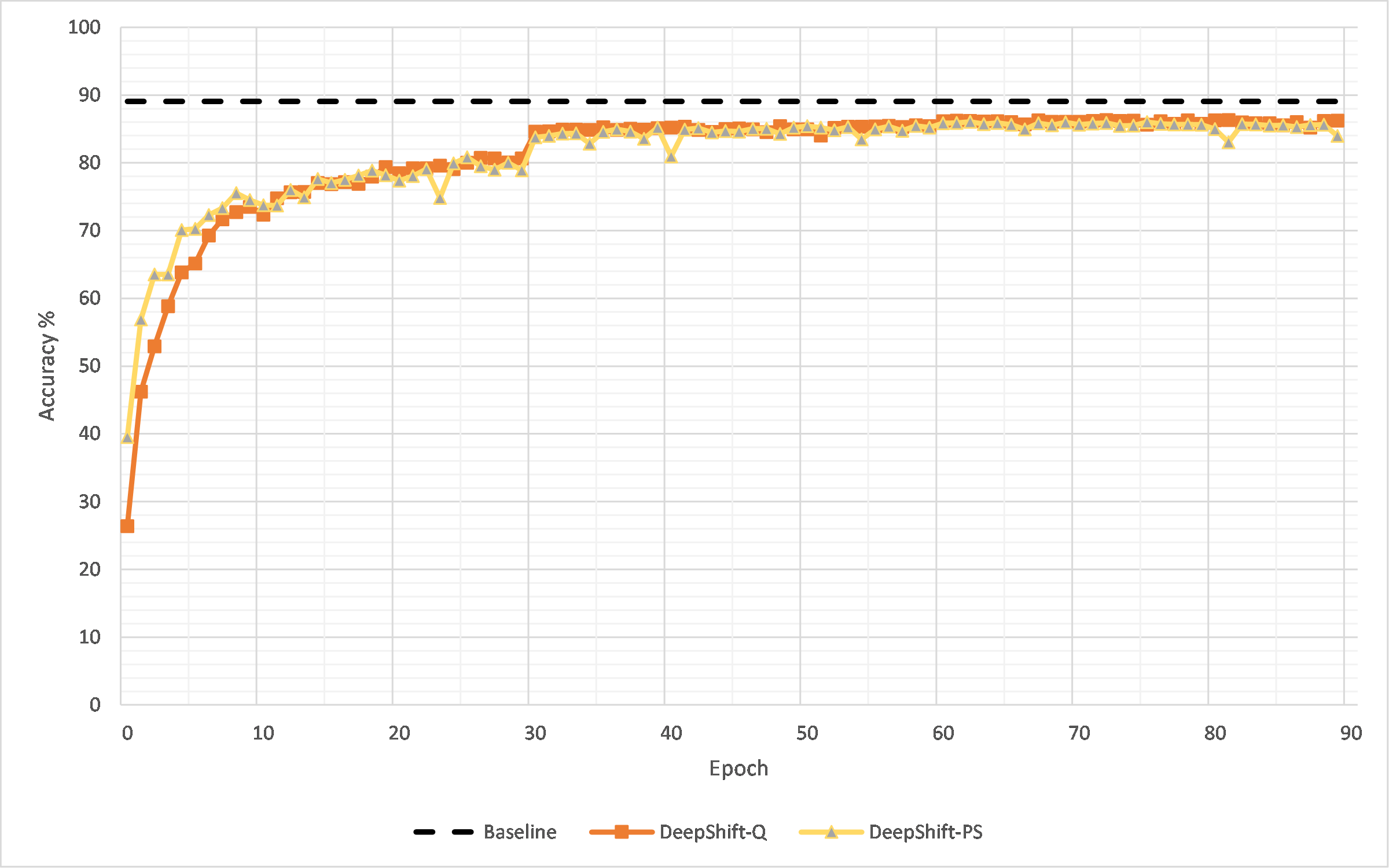}
		\caption{ResNet18 Top-5 Accuracy}\label{fig:imagenet:resnet18-top5}
	\end{subfigure}
	\begin{subfigure}[b]{.4\textwidth}
	\includegraphics[width=\textwidth]{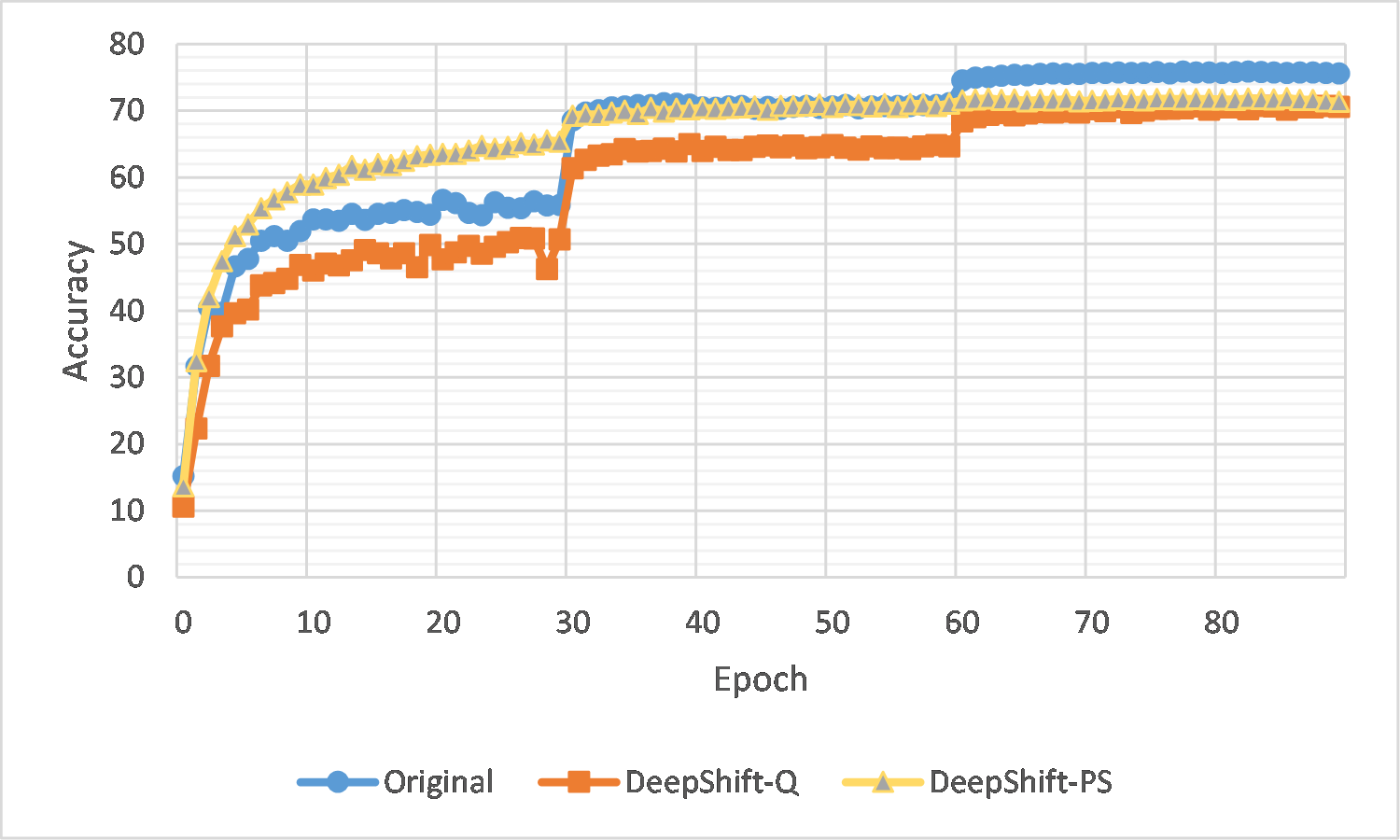}
	\caption{ResNet50 Top-1 Accuracy}\label{fig:imagenet:resnet50-top1}
	\end{subfigure}\qquad
	\fi
	\begin{subfigure}[b]{.4\textwidth}
		\includegraphics[width=\textwidth]{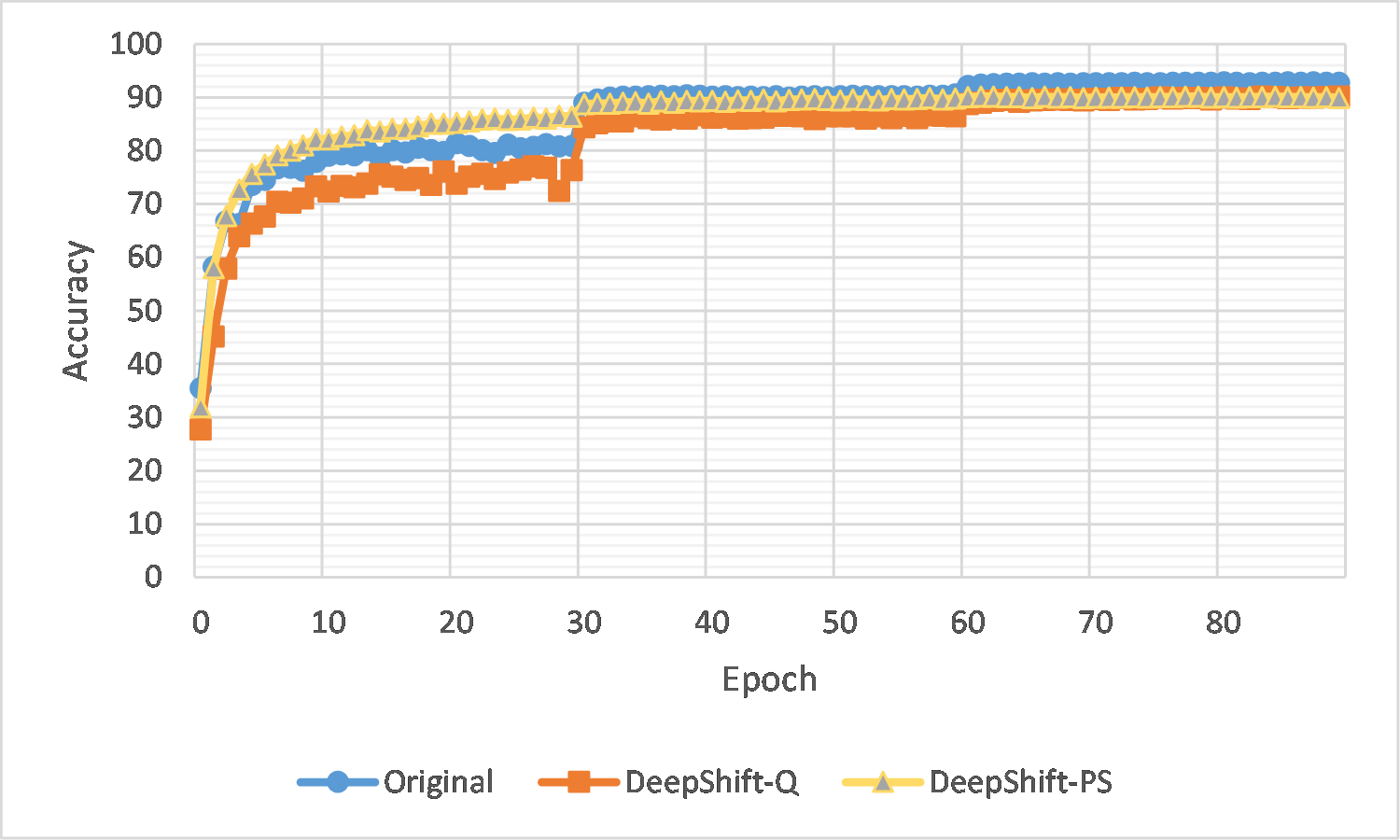}
		\caption{ResNet50 Top-5 Accuracy}\label{fig:imagenet:resnet50-top5}
	\end{subfigure}
	\ifdefined\arxiv
	\begin{subfigure}[b]{.4\textwidth}
		\includegraphics[width=\textwidth]{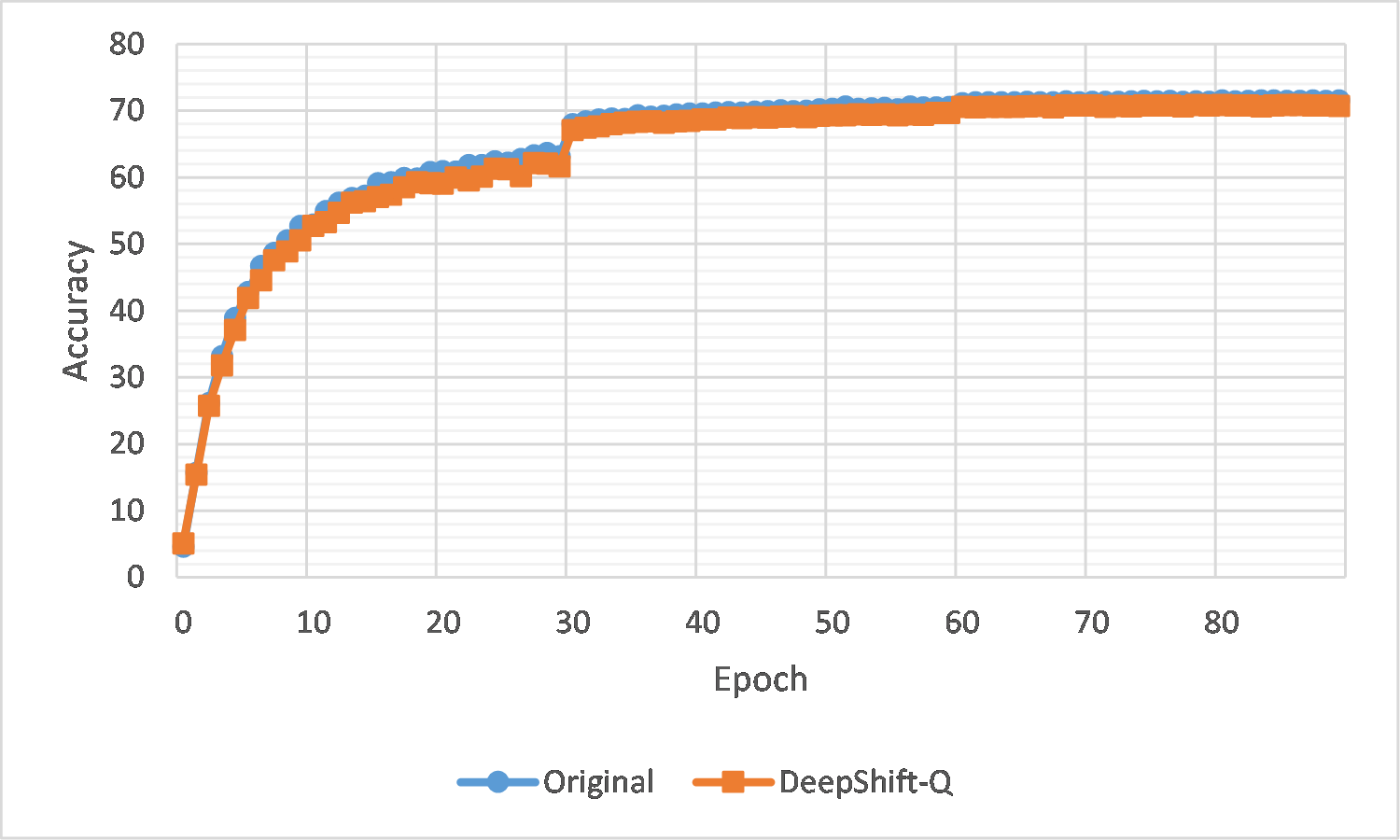}
		\caption{VGG16 Top-1 Accuracy}\label{fig:imagenet:vgg16-top1}
	\end{subfigure}
	\fi
	\begin{subfigure}[b]{.4\textwidth}
		\includegraphics[width=\textwidth]{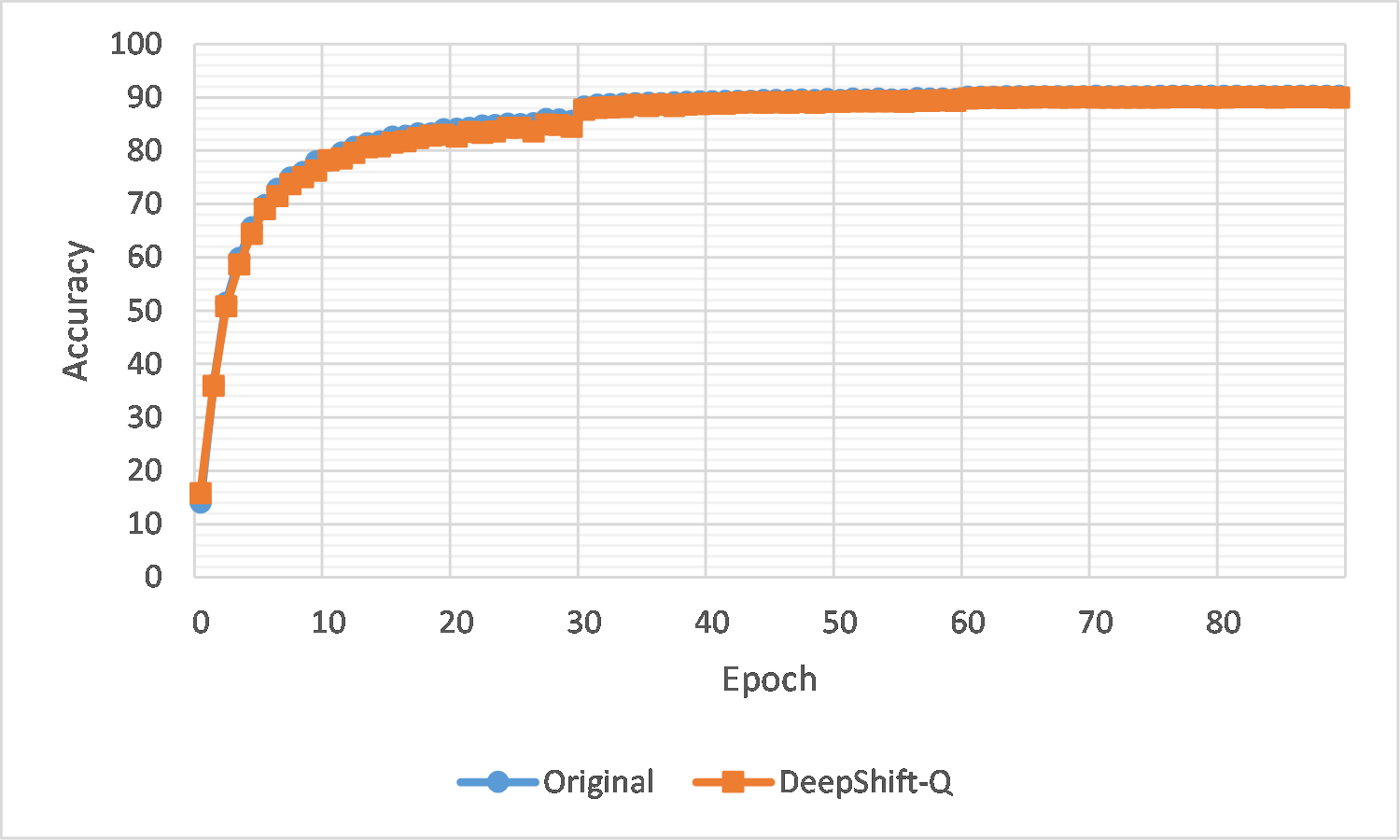}
		\caption{VGG16 Top-5 Accuracy}\label{fig:imagenet:vgg16-top5}
	\end{subfigure}
	\caption{Imagenet training from scratch}
	\label{fig:imagenet}
\end{figure*}

Models were trained from scratch for 90 epochs, the learning rate used to train from scratch was 0.01 that decays by a factor of 0.1 every 30 epochs. Training plots are shown in Figure \ref{fig:imagenet} and results are shown in Table \ref{imagenet-results-from-scratch}. 

For ResNet18, we compare the performance of DeepShift that replaces multiplications with bitwise shifts and sign flips, with BWN (Binary Weight Networks) \& TWN (Ternary Weight Networks) that replace multiplications with sign flips only.

\begin{table*}
	\caption{Imagenet Accuracy Results: Training from Scratch. ``Compute'' refers to the type of arithmetic instruction the convolution and fully-connected layers are based on. }
	\label{imagenet-results-from-scratch}
	\centering
	\begin{tabularx}{\textwidth}{lcc*{4}{>{\centering\arraybackslash}X}}
		\toprule
		Model & Method & Compute & \multicolumn{2}{c}{Top-1} & \multicolumn{2}{c}{Top-5} \\ 
		\cmidrule(r){4-7}
		& & & Accuracy & Delta & Accuracy & Delta \\
		\midrule
		ResNet18 & Original & Multiply & 69.76\%  & - & 89.08\% & - \\
		& DeepShift-Q [Ours] & Shift \& Sign Flip & 65.32\% & -4.44\% & 86.30\% & -2.78\%  \\
		& DeepShift-PS [Ours] & Shift \& Sign Flip & 65.34\% & -4.42\% & 86.05\% & -3.03\%  \\
		& BWN* \cite{Rastegari2016} & Sign Flip & 60.80\% & -9.30\% & 83.00\% & -6.20\% \\
		& TWN* \cite{Li2016} & Sign Flip & 61.80\% & -3.60\% & 84.20\% & -2.56\% \\ 
		\midrule
		\bottomrule
		ResNet50 & Original & Multiply & 76.13\% & - & 92.86\% & - \\
		& DeepShift-Q [Ours] & Shift \& Sign Flip & 70.73\% & -5.56\% & 90.16\% & -2.70\%  \\
		& DeepShift-PS [Ours] & Shift \& Sign Flip & 71.90\% & -4.23\% & 90.16\% & -2.70\%  \\
		\midrule
		\bottomrule
		VGG16 & Original & Multiply & 71.59\% & - & 90.38\% & - \\
		& DeepShift-Q [Ours] & Shift \& Sign Flip & 70.87\% & -0.71\% & 90.09\% & -0.29\%  \\
		\midrule
		\bottomrule
	\end{tabularx}
\end{table*}

\section{Efficiency Analysis} \label{sec:efficiency-analysis}
Multiplication in integer or fixed-point format consists of multiple bitwise shifts, ANDs, and additions. Floating point (FP) multiplication consists of more steps such as adding the exponents, \& normalizing. In terms of CPU cycles, taking a 32-bit Intel Atom instruction processor as an example, integer and FP multiplication instruction take 5 to 6 cycles, while bit-wise shift takes only 1 cycle (\cite{AgnerInstructionTables}). (\cite{Asati2009}) shows for 16-bit architectures, average power, and area of multipliers are at least $9.7\times$, and $1.45\times$, respectively, of bitwise shifter: 22.05 nW to 54.83 nW vs 1.32 nW to 2.27 nW, and 8.7K to 29K transistor count vs 2K to 6K only. A common optimization in C++ compilers is to detect an integer multiplication with a power of 2 and replace it with a bitwise shift. Our contribution is enabling bitwise shifts in training neural networks while achieving good accuracy. We emphasize that we replace each multiplication with a single bitwise shift. \\ 

DeepShift also achieves model storage compression by representing the weights with fewer bits. This reduces the memory footprint of the model and increases energy savings for transferring weights between different layers of memory of a GPU or CPU (\cite{Sze_2017}). That is needed in mobile and IoT applications, where memory footprint is a primary constraint. Energy savings is crucial for applications like drones that fly on batteries, whether the computing platform is a low-end CPU, GPU, or FPGA. Also, bitwise shift operators are a good candidate for in-memory computing research \cite{Sze_2017}, as they're cheaper in size and energy to place in or near memory than multiplication.

\ifdefined\arxiv
\subsection{GPU Implementation} \label{subsec:gpu-implementation}
\ifdefined\arxiv
While custom hardware designs have been proposed by \cite{LogNet_7953288}, \cite{Vogel:2018:EHA:3240765.3240803} and \cite{ShiftCNN}, we have implemented a GPU kernel that obtained a speedup of  26\% on ResNet18 on ImageNet. To the best of our knowledge, this is the first paper to implement a CUDA kernel convolution using bitwise shifts.

39\% on the CIFAR10 version of ResNet18 and a speedup of
The total time to infer the validation images of each dataset is shown in Table \ref{table:speedup}.
\fi
\begin{table*}[!t]
	\caption{Inference Time Speed Up of DeepShift. Inference time shown is the total time to infer the validation images of the corresponding dataset.}
	\label{table:speedup}
	\centering
	\begin{tabular}{lllll}
		\toprule
		Model & Dataset & \multicolumn{2}{l}{Inference Time} & Inference Speed Up \\
		\cmidrule(r){3-4}
		&                          & Original        & DeepShift        &                                     \\
		\midrule 
		ResNet18               & CIFAR10                  & 19.42 s         & 11.80 s          & 39\%                                \\
		ResNet18               & ImageNet                 & 2010 s          & 1490 s           & 26\%    \\
		\bottomrule                           
	\end{tabular}
\end{table*}

It is noteworthy that the baseline that we compared against is a custom multiplication-based convolution kernel which is slower than NVIDIA's cuDNN optimized kernel. Future work can be done to optimize our bitwise-shift CUDA kernels to compare them with NVIDIA's cuDNN multiplication-based kernels, such as fusing convolution with ReLu, tuning the tiling factors, and performing JIT (just-in-time) compilation of CUDA kernels at runtime.

 Some of the challenges that we faced during implementation that limited the speedup of the kernel: were that the GPU's instruction set architecture was not optimized for using smaller bit-width parameters that were needed to represent the shift and sign. Having to extract the 4-bit shift value and 1-bit sign value from 32-bit integers using masking and extra shifts incurred a lot of overhead. Also, if-else statements on the values of the sign bits to determine whether to add or subtract during accumulation or if-else conditions on whether shift values are positive or negative to determine whether to shift left or right, adding more overhead.

The implementation of the kernel can be found our GitHub repo. 

\fi




\section{Future Work}
\ifdefined\arxiv
We have proven that we can train models using bitwise shift and sign flips from pre-trained full-precision models and obtain accuracies above or almost the same as the original accuracies. This can minimize energy and latency for on-device training for applications like transfer learning on the edge. Although training from scratch resulted in drops of 2\% to 3\% in Top-5 accuracy, DeepShift can also save energy consumption on the cloud.

The hardware realization of DeepShift neural networks is yet to be done and is needed to evaluate the actual speedup in performance. Designing parallel architectures for bitwise shift and negation on vectors rather than on individual registers may face its own challenges but it is expected to have faster execution times than their multiplication counterparts. 

While other methods that tackle neural network speed-ups such as BNNs perform well on small datasets (e.g., MNIST and CIFAR10) but suffer significant degradation on big datasets such as Imagenet, DeepShift networks proved that they are suitable for Imagenet.

Also, a GPU kernel of performing convolution using bitwise shift has been implemented as a proof-of-concept. Developing GPU instruction set architectures to make them more optimized for smaller bit-width data types, and shift operations (e.g., an assembly instruction that can deduce bitwise shift should be done in the opposite direction if the shift value is negative) will help in making bitwise-shift-based training and inference more common. Moreover, the existence of vector CPU instructions that can perform bitwise shifts given a vector of integers, and a vector of shift values, can make DeepShift more optimal on CPUs.
\fi
Evaluating DeepShift on custom FPGA design, or extending CPU and GPU instruction sets with dedicated bitwise shift and negate instructions that operate on a vector and/or compressed 5-bit representation is one direction to explore. Also, we suggest training models from scratch with both activations and weights represented as powers of 2, similar to \cite{Miyashita2016ConvolutionalNN} that did that for inference only. Moreover, to speed up training, we suggest to try gradients represented as 8-bit or 16-bit fixed points.

\subsubsection*{Acknowledgments}

We thank Ahmed Eltantawy for providing us valuable feedback to implement the CUDA kernels, and Sara Elkerdawy for providing valuable insights during our discussions. \ifdefined\arxiv We thank also the PyTorch team for providing example training scripts and pre-trained model files for the Imagenet dataset that allowed us to reproduce the accuracies of the original models' papers. \fi

{\small
\bibliographystyle{ieee_fullname}
\bibliography{paper}
}

\end{document}